\documentclass[10pt,journal,compsoc]{IEEEtran}

\usepackage{xcolor}
\usepackage{multirow}
\usepackage{amsmath}
\usepackage{comment}
\usepackage{soul,color}
\usepackage[gen]{eurosym}
\usepackage{hyperref}
\usepackage{enumitem}
\setlist[itemize]{noitemsep, topsep=0pt}
\usepackage{amssymb}
\usepackage{graphicx}
\usepackage{caption}
\usepackage{comment}
\usepackage{subcaption}
\usepackage[utf8]{inputenc}
\usepackage[linesnumbered,ruled,vlined,algo2e]{algorithm2e}
\usepackage{multirow}
\usepackage{subcaption}
\AtBeginDocument{%
  \providecommand\BibTeX{{%
    \normalfont B\kern-0.5em{\scshape i\kern-0.25em b}\kern-0.8em\TeX}}}
\usepackage{balance }

\begin{document}
\def\x{{\mathbf x}}
\def\L{{\cal L}}
\def\eg{\textit{e.g.}}
\def\ie{\textit{i.e.}}
\def\Eg{\textit{E.g.}}
\def\etal{\textit{et al.}}
\def\etc{\textit{etc}}
\markboth{Journal of \LaTeX\ Class Files,~Vol.~14, No.~8, August~2024}%
{Shell \MakeLowercase{\textit{et al.}}: On the Validity of Head Motion Patterns as Generalisable Depression Biomarkers}

\title{On the Validity of Head Motion Patterns as Generalisable Depression Biomarkers}
\author{Monika~Gahalawat,~\IEEEmembership{Student Member, IEEE}, Maneesh~Bilalpur,~\IEEEmembership{Student Member, IEEE}, Raul~Fernandez~Rojas,~\IEEEmembership{Member, IEEE}, Jeffrey F. Cohn,~\IEEEmembership{Senior Member, IEEE}, Roland Goecke,~\IEEEmembership{Senior Member, IEEE} and, ~Ramanathan~Subramanian,~\IEEEmembership{Senior Member, IEEE}}

\IEEEtitleabstractindextext{%
\begin{abstract}
   Depression is a debilitating mood disorder negatively impacting millions worldwide. While researchers have explored multiple verbal and non-verbal behavioural cues for automated depression assessment, \textit{\textbf{head motion}} has received little attention thus far. Further, the common practice of validating machine learning models via a single dataset can limit model generalisability.
   This work examines the effectiveness and generalisability of models utilising elementary head motion units, termed \emph{kinemes}, for depression severity estimation. Specifically, we consider three depression datasets from different western cultures (German: \emph{AVEC2013}, Australian: \emph{Blackdog} and American: \emph{Pitt} datasets) with varied contextual and recording settings to investigate the generalisability of the derived kineme patterns via two methods: (i) \textit{k-fold} cross-validation over individual/multiple datasets, and (ii) model reuse on other datasets. Evaluating classification and regression performance with classical machine learning methods, our results show that: (1) head motion patterns are efficient biomarkers for estimating depression severity, achieving highly competitive performance for both classification and regression tasks on a variety of datasets, including achieving the second best Mean Absolute Error (MAE) on the AVEC2013 dataset, and (2) kineme-based features are more generalisable than (a) raw head motion descriptors for binary severity classification, and (b) other visual behavioural cues for severity estimation (regression).  
\end{abstract}

}


\maketitle
\section{Introduction}\label{Sec:Intro}
Depression is a highly prevalent mood disorder~\cite{institute2021global} characterised by a prolonged (typically more than two weeks) feeling of sadness, emptiness, and a lack of energy accompanied by cognitive and somatic changes. These changes negatively influence an individual’s ability to function in daily life. Depression is often linked to an increase in comorbidities including anxiety and substance abuse disorders, hypertensive diseases and diabetes, possibly leading affected individuals to suicide~\cite{goldney2000suicidal}. It exerts a significant economic burden globally incurring direct healthcare costs~\cite{greenberg2015economic}. The situation is further exacerbated by the indirect costs through lost labour force participation and productivity~\cite{schofield2019indirect}, highlighting the critical need for early diagnosis and intervention. Although depressive disorders can be effectively treated, the current diagnosis process relies heavily on self-reports and clinical observations, resulting in a lack of objectivity and high susceptibility to perceptual biases~\cite{crapanzano2018exploration}. 

Numerous psychology and affective computing studies have focused on developing objective measures to estimate depression severity over the last decade \cite{cohn2018multimodal, pampouchidou_et_al_TAC_DepressionReview}. Early depression detection studies focused on leveraging non-verbal behavioural expressions such as facial movements~\cite{bourke2010processing}, eye gaze~\cite{alghowinem2016multimodal}, body gestures~\cite{joshi2013relative},  head movements~\cite{alghowinem2013head} and speech features \cite{cummins2011investigation}. Recent approaches have estimated severity of depression from multiple modalities employing deep learning methods~\cite{uddin2022deep, fang2023multimodal}. The availability of several audiovisual datasets, primarily from western cultures, such as the Audio/Visual Emotion Challenge (AVEC)~\cite{valstar2013avec}, Black Dog Institute dataset~\cite{alghowinem2013head}, University of Pittsburgh  dataset~\cite{yang2012detecting}, and Distress Analysis Interview Corpus -- Wizard of Oz~\cite{gratch2014distress} has been instrumental in driving this change. 

Despite multiple datasets available to aid depression benchmarking, most studies validate their models on a solitary dataset leading to models lacking robustness and generalisability to broader, culturally diverse datasets. Especially, the prevalence of varied depression symptom profiles across cultures and ethnicities~\cite{gabriella2012cultural, marsella2003cultural} necessitates the development of generalisable depression detection techniques.    
\begin{figure*}[t]
      \centering
     \includegraphics[width=0.9\linewidth]{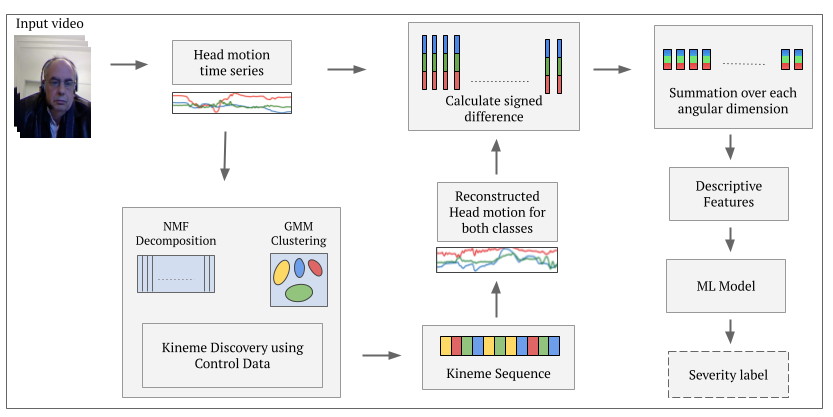} \vspace{-3mm}
    \caption{\textbf{Overview:}  \textmd{Kineme patterns are solely learnt from \textit{control} subjects, and the reconstruction error is computed between the actual vs.\ reconstructed head-motion segments for both the \textit{control} and \textit{depressed} classes. Statistical descriptors over the \textit{yaw}, \textit{pitch} and \textit{roll} angular dimensions ($8 \times 3$ features) are utilised for depression severity estimation.~\cite{gahalawat2023explainable} } } 
    \label{fig:Depression_proposed_framework}
\end{figure*}
Building on preliminary findings on depression detection using elementary head motion or~\emph{kinemes}~\cite{gahalawat2023explainable}, this work examines the utility of kinemes for estimating depression severity on datasets reflecting three different western cultures, and compiled under different recording conditions, participant tasks, and depression measures. We explore cross-corpus generalisability over the: 

\begin{itemize}
    \item \textbf{AVEC2013 (AVEC) dataset} is a subset of Audio-Visual Depressive language corpus (AViD) with videos of subjects performing different Powerpoint-guided tasks. This German dataset provides participant self-reported scores based on the Beck Depression Index (BDI).
    \item \textbf{University of Pittsburgh (Pitt) dataset} monitors depression over multiple weeks via interviews with Euro or African-American participants based on the Hamilton Rating Scale for Depression (HRSD) questionnaire.
    \item \textbf{Black Dog Institute (Blackdog) depression dataset} contains structured interview responses from both healthy and depressed cohorts. This dataset compiled in Australia is clinically validated and contains the Quick Inventory of Depressive Symptomatology-Self Report (QIDS-SR) values.
\end{itemize}
%

Prior studies have focused on capturing the dynamics of head movement by measuring the variation in amplitude, velocity and acceleration of 3D head pose angles from all classes. However, we discover the kineme patterns solely from the head movements of healthy control (or low-depressed) individuals, and employ those to reconstruct movements for both the \textit{healthy} and \textit{depressed} classes. Additionally, in contrast to previous studies, the statistical features are calculated based on the reconstruction errors between the actual and derived head movement vectors for the two classes. Several classification and regression models are trained on these statistical features to demonstrate the effectiveness of kineme features over the three datasets, and to examine the generalisability of these patterns. Overall, our research contributions are summarised below: 
%
\begin{enumerate}
    \item This study is the first to illustrate the effectiveness of kinemes for depression severity estimation using an ordinal measure rather than dichotomous discrimination.
    \item On the AVEC test set, an MAE/RMSE score of 5.68/7.57 is achieved, outperforming prior head pose-based approaches, and performing comparably to state-of-the-art (SOTA) visual methods.
    \item The experiments demonstrate greater generalisability of kinemes compared to a) raw head pose features for classification and b) facial movements for regression. 
    \item Examining individual datasets, kineme patterns discovered from AVEC show greater generalisation power than kinemes learned from BlackDog and Pitt. 
\end{enumerate}
The remainder of this paper is organised as follows. Section~\ref{Sec:RW} reviews depression analysis using head movements, and cross-corpus generalisability. Section~\ref{Sec:Meth} details kineme formulation and statistical feature extraction, while Section~\ref{sec:Datasets} describes datasets used in this study. Section~\ref{Sec:ER} summarises aspects considered to manage variability across datasets, and details the performance measures and models implemented. 
Section~\ref{sec:RnD}  discusses classification and regression results, while conclusions are drawn in Section~\ref{Sec:Con}.
%
%

\section{Related Work}\label{Sec:RW}
We now review relevant literature on (a) depression assessment using head motion, and (b) cross-corpus generalisability of depression detection models.

\subsection{Depression Analysis using Head Motion}
Research on depression assessment using behavioural cues~\cite{waxer1974nonverbal, pedersen1988ethological} has shown that head motion is an effective biomarker.  Waxer~\etal~\cite{waxer1974nonverbal} identified head pose as a major depression cue as depressed subjects were more likely to keep their head in a downward position. Reduced head nodding was observed in depressed individuals by Fossi~\etal~\cite{fossi1984ethological}. Another study~\cite{pedersen1988ethological} highlighted  pronounced behavioural changes in the head and hand regions for depressed patients. Recent studies illustrate a decrease in overall movements due to depression symptoms~\cite{mitsue2019relationship}, and lower head motion exhibited by depressed patients~\cite{horigome2020evaluating}.

Over the last decade, objective methods for depression benchmarking have been developed leveraging verbal (\cite{cummins2011investigation, huang2019investigation}) and non-verbal (\cite{bourke2010processing, joshi2013relative}) behavioural cues. However, most studies focus on facial dynamics or body gestures with head movements receiving little attention. Alghowinem~\etal~\cite{alghowinem2013head} used head pose cues for depression detection and Joshi~\etal~\cite{joshi2013can} discovered fewer head movements in depressed subjects with a higher frequency of static head positions. Kacem~\etal~\cite{kacem2018detecting} encoded head motion dynamics to benchmark depression severity as\textit{ mild}, \textit{moderate} or \textit{severe}. Some studies have utilised head motion as a complementary cue to verbal features and eye gaze~\cite{alghowinem2016multimodal, alghowinem2020interpretation}. 
%

\subsection{Cross-corpus Generalisability}
Due to the ethical, clinical and legal constraints relating to collecting and sharing depression datasets, very few studies analyse cross-corpus generalisability. Most researchers validate their models on solitary datasets leading to models lacking generalisability and robustness. Additionally, differences in depression symptom profiles owing to national, ethnical and cultural factors~\cite{gabriella2012cultural, marsella2003cultural} necessitate model analysis using diverse datasets. Alghowinem~\etal~\cite{alghowinem2015cross} investigated generalisability of a classifier trained via eye activity and head movement features across three cross-cultural depression datasets. Another study~\cite{alghowinem2020interpretation} explored generalisability of features extracted from eye movement, head pose, and speech prosody. Ahmad~\etal~\cite{ahmad2021cnn} examined the generalisability of a CNN model for depression severity estimation over two datasets.
%

\subsection{Analysis of related literature}
A review of the literature conveys that (i) while head pose movements are utilised as a complementary cue, very few works have exclusively explored head motion for depression severity estimation, (ii) little research has examined the utility and generalisability of non-verbal behavioural cues for depression. To address this research gap, this study (a) examines the performance of head motion-based kinemes for estimating depression severity, (b) presents a detailed investigation of cross-corpus generalisability using kinemes for depression classification and regression. Our results show the efficacy of kinemes for depression estimation, and demonstrate the better generalisability of kinemes in comparison to raw head pose and facial features. 

%
%
\section{Methodology}\label{Sec:Meth}
This section outlines the approach implemented to discover kineme patterns from short overlapping head pose segments, and compute statistical features for model training.   
%

%
\subsection{Kineme Formulation}\label{Sec:KF}
For each video, 3D head pose angles (\emph{pitch}, \emph{yaw} and \emph{roll}) are extracted using the OpenFace~\cite{Baltrusaitis16} toolkit for facial detection and head pose estimation. We represent head pose angles as a time-series of short overlapping segments to enable shift invariance. These segments are projected to a lower dimensional space via Non-negative Matrix Factorisation (NMF) and clustered using a Gaussian Mixture Model (GMM)  approach in the learned subspace~\cite{samanta2017role}. 

We express the 3D Euler rotation angles, \emph{pitch} ($\theta_p$), \emph{yaw} ($\theta_y$) and \emph{roll} ($\theta_r$) over a duration $T$ as a time-series: $\boldsymbol{\theta} = \{\theta_p^{1:T}, \theta_y^{1:T}, \theta_r^{1:T}\}$. To ensure non-negativity, head pose angles are maintained in [0$^{\circ}$, 360$^{\circ}$]. Each individual segment from time-series $\theta$ is represented as a vector, where the $i^{th}$ segment is expressed as $\mathbf{h}^{(i)} = [\theta_p^{i:i+\ell}\, \theta_y^{i:i+\ell}\, \theta_r^{i:i+\ell}]$. All head pose segments have a uniform length $\ell$ with an overlap of $\ell/2$. Assuming that a particular video contains a total of $s$ segments, the characterisation matrix $\mathbf{H}_{\boldsymbol\theta}$ is defined as $ \mathbf{H}_{\boldsymbol\theta} = [\mathbf{h}^{(1)}, \mathbf{h}^{(2)},\cdots, \mathbf{h}^{(s)}]$. For a training dataset of $N$ videos, the $\mathbf{H}_{\boldsymbol\theta}$ matrices of individual videos are concatenated to form the head motion matrix, $\mathbf{H} = [\mathbf{H}_{\boldsymbol\theta_1}|\mathbf{H}_{\boldsymbol\theta_2}|\cdots|\mathbf{H}_{\boldsymbol\theta_N}]$, where each column represents a head pose segment. Subsequently, NMF is implemented to decompose the head-motion matrix $\mathbf{H}\in\mathbb{R}_+^{m\times n}$ into a basis $\mathbf{B}\in\mathbb{R}_+^{m\times q}$ and a coefficient matrix $\mathbf{C}\in\mathbb{R}_+^{q\times n}$ with $m = 3\ell$, $n = Ns$ as follows:
%
\begin{equation}
    \underset{\mathbf{B} \geq 0, \mathbf{C} \geq 0}{\text{ min}} \lVert{\mathbf{H} - \mathbf{B}\mathbf{C}}\rVert_F^2
\end{equation}
where $q \leq min(m, n)$ and $\lVert \textbf{ . }  \rVert_F$ denotes the Frobenius norm. We build a GMM to categorise columns of the coefficient matrix $\mathbf{C}$ to produce a ${\mathbf{C}^*}\in\mathbb{R}_+^{q\times k}$ where $k << Ns$ in the learned subspace. Finally, the centers of these GMM clusters are transformed back to the original subspace using $\mathbf{H}^*=\mathbf{B}\mathbf{C}^*$, where the learned set of $k$ kinemes $\{\mathcal{K}_i\}_{i=1}^k$ are denoted by the columns of the matrix $\mathbf{H}^*$.  
Utilising the learned set of kinemes $K$, any head motion time-series $\theta$ can be expressed as a kineme sequence by associating each head pose segment with the closest kineme. To compute the kineme label for the $i^{th}$ segment, the characterisation vector $\mathbf{h}^{(i)}$ is projected onto the learned subspace defined by $\mathbf{B}$ to yield $\mathbf{c}^{(i)}$ such that:
\begin{equation}
    \hat{\mathbf{c}} = \underset{\mathbf{c}^{(i)} \geq 0}{\text{arg min}} \lVert{\mathbf{h}^{(i)} - \mathbf{B}\mathbf{c}^{(i)}}\rVert_F^2
\end{equation}
We map the $i^{th}$ segment of the time-series with its corresponding kineme $K^{(i)}$ by maximising the posterior probability $P({K}|\hat{\mathbf{c}})$ over all kinemes to obtain the kineme sequence $\{K^{(1)} \cdots K^{(s)}\}$ , where $K^{(j)}\in \mathcal{K}$. In this study, each head motion time-series is created by sampling 10 frames per second (fps) for the initial 5 minutes to maintain consistency among the three datasets. We consider a higher sampling rate for videos shorter than 5 minutes to obtain uniform-length time-series. The series is divided into short segments of $\ell$ = 5s each with an overlap of 2.5s. A total of $k$ = 16 kinemes are discovered from the \textit{healthy} cohort data. 

%

%
\subsection{Feature Extraction}\label{Sec:FE}
While kineme patterns are discovered from \textit{healthy} (or low depressed)  cohort, both \textit{healthy} and \textit{depressed} classes are then represented in terms of the learned kinemes. We computed the reconstruction error between the raw and learned head pose measures over each angular dimension, and derived a set of eight descriptive features therefrom. To compute reconstruction error, suppose the head pose vector $\mathbf{h}^{(i)}$ for the $i^{th}$ segment is associated with the kineme value $K^{(i)}$. Each individual $K^{(i)}$ can be expressed by head pose values as $\tilde{\mathbf{h}}^{(i)}$, determined by converting the GMM cluster centre values to the original \emph{pitch}-\emph{yaw}-\emph{roll} subspace. If the reconstructed head pose vector for the $i^{th}$ segment is denoted as:  
\begin{equation}
    \tilde{\mathbf{h}}^{(i)} = [\tilde{\theta}_p^{i:i+\ell}\, \tilde{\theta}_y^{i:i+\ell}\, \tilde{\theta}_r^{i:i+\ell}]
\end{equation}
The reconstruction error is then computed as the signed difference between the actual head pose vector and the learned GMM cluster centres. For each individual segment, the difference vector $\mathbf{d}^{(i)}$ can be computed as: 
\begin{equation}
    \mathbf{d}^{(i)} = \mathbf{h}^{(i)} - \tilde{\mathbf{h}}^{(i)} = [d_p^{i:i+\ell}\, d_y^{i:i+\ell}\, d_r^{i:i+\ell}]
\end{equation}
We summed the signed differences over the pitch (\emph{p}), yaw (\emph{y}), and roll (\emph{r}) dimensions  and over all segments to compute a set of descriptive features. The aggregate of these signed differences over the $i^{th}$ segment is expressed as
\begin{equation}
        {s_e}^{(i)} = \sum_{n=1}^{\ell} d_e^{i:i+n} 
\end{equation}
where each ${s_e}^{(i)}$ is calculated for the three angular dimensions $e \in \{p, y, r\}$ over all head pose segments.
Considering a specific chunk size $n_c$ for classification/regression, we obtain the feature vector per angle $e \in \{p, y, r\}$ as
\begin{equation}
        \mathbf{as_e} = [\lvert{s_e}^{(1)} \rvert, \lvert{s_e}^{(2)}\rvert, \cdots, \lvert{s_e}^{(n_c)}\rvert ] 
\end{equation}
with $\lvert \cdot \rvert$ denoting the absolute value. Based on~\cite{gahalawat2023explainable}, these feature vectors are then utilised to compute eight statistical features, namely, \emph{minimum}, \emph{maximum}, \emph{range}, \emph{mean}, \emph{median}, \emph{standard deviation}, \emph{skewness}, and \emph{kurtosis} over the three dimensions to obtain a total of $8 \times 3$ features for each chunk.  
%
%

\section{Datasets}\label{sec:Datasets}
We considered three datasets compiled from different western cultures for evaluating generalisability. 

\textbf{AVEC~\cite{valstar2013avec}:} is a subset of the audio-visual depressive language corpus (AViD-corpus) that contains 340 video recordings of German subjects aged 18--63 yr performing different PowerPoint guided tasks such as counting 1--10, talking about a happy/sad experience, reading text, and talking out loud during task performance. AVEC contains 150 videos split into train, development and test categories with videos of length 20--50 min. Only one person is present in the video frame, although several subjects appear multiple times in this dataset. All participants completed the self-reported Beck-Depression Inventory (BDI)~\cite{beck1996comparison}, a 21-question multiple-choice self-report inventory, with scores between 0--63 denoting  depression levels~\cite{valstar2013avec}. For binary classification, subjects with a BDI score $\leq$ 13 are categorised as \textit{low-depressed}, with others deemed \textit{severely depressed}.

\textbf{Pitt dataset~\cite{yang2012detecting}:} contains interview data from 49 Euro- or Afro-American depressed participants selected from a clinical trial for depression treatment. Hamilton Rating Scale for Depression (HRSD) based questionnaire was used to conduct interviews by probing the participants’ mood, and feelings of guilt, suicide ideation, and anxiety. These interviews were recorded using three cameras and two unidirectional microphones with $640 \times 480$ pixel video resolution, and audio digitised at 48 kHz. The videos were labeled as per the clinical-rated HRSD measure with each point rated on a 3 or 5-point Likert scale. A HRSD score of 7 or lower indicates \textit{remission}, 15 or higher depicts \textit{moderate-to-severe depression}, and any score from 8--14 indicates \textit{mild depression}. For binary classification, we categorised a score of $\leq$ 7 as \textit{low-depressed}, and others as \textit{severely depressed}.

\textbf{Blackdog Dataset~\cite{alghowinem2016multimodal}:} was collected at the Black Dog Institute, Sydney, Australia. Clinicians gave careful considerations to participant selection -- depressed patients were selected based on the Diagnostic and Statistic Manual of Mental Disorders (DSM-IV) criteria, while the healthy controls comprised participants with an IQ$>80$ and no history of drug addiction or mental illness. Here, we consider only the structured interview task where 90 subjects responded to a clinician’s open-ended questions designed to elicit spontaneous and emotional responses. Participants are categorised into the \textit{healthy control} and \textit{depressed patient} classes.
%

\begin{table*}[ht]
\renewcommand{\arraystretch}{1.2}
\fontsize{7}{7}\selectfont
\caption{Overview of the datasets used in this study.}
\vspace{-2mm}
\centering
\begin{tabular}{|l||c|c|c|} 
\hline
    \bf Dataset    & \textbf{AVEC} & \textbf{Pitt} & \textbf{Blackdog} \\ 
    \hline \hline
    \textbf{Dataset Language} &  German & English (American) & English (Australian) \\ 
    \textbf{Number of participants} & 82 (German) & 49 (Euro- and Afro-American) & 90 (Australian)  \\ 
    \textbf{Number of videos} & 150 & 148 & 90  \\
    \textbf{Age Distribution} & 18--63 years & 19--65 years & 21--75 years  \\ 
    \textbf{Procedure}& Human-computer interaction experiment & HRSD clinical interview & Open-ended questions interview  \\ 
    \textbf{Classification}& Severe/Low depressed & Severe/Low depressed & Severely depressed/Healthy controls  \\ 
    \textbf{Severity measure} & BDI & HRSD & QIDS-SR  \\ 
    \textbf{Hardware} & 1 web camera + 1 microphone & 4 cameras + 2 microphones & 1 camera + 1 microphone \\ 
    \textbf{Video sampling rate}  & 30 fps & 30 fps & 30 fps  \\ 
    \textbf{Audio sampling rate} & 44100 Hz & 48000 Hz & 44100 Hz  \\ 
    \hline
\end{tabular} 
\label{tab:Dataset_desc}
\end{table*}


%
Table~\ref{tab:Dataset_desc} presents an overview of the datasets along with the details of participants considered in our study.  For regression analysis, we convert the severity labels for AVEC and Pitt dataset to QIDS-SR values, similar to Blackdog. The Inventory of Depressive Symptomatology (IDS) and the QIDS conversion table are utilised to convert the HRSD and BDI severity scales to \href{https://web.archive.org/web/20191101195225/http://www.ids-qids.org:80/interpretation.html}{QIDS-SR}.  


%
%
\section{Experiments}\label{Sec:ER}

%
\subsection{Preprocessing}\label{Sec:Pre}
As recording conditions varied across datasets, we cropped all video frames using facial detection to eliminate visual information that could hamper inference. This step facilitates accurate and consistent estimation of head movements using the \textit{Openface} toolkit. The \textit{Dlib} \textit{OpenCV} module was used for face detection, based on which each frame was cropped and a padding margin included to compensate for potential detection errors and excessive head movements.  
%

%
\subsection{Kineme Discovery}\label{Sec:KD} 
As stated in Sec.~\ref{Sec:KF}, 16 kinemes were learned based on head pose segments of 5s length and an overlap of 2.5s. To examine generalisability of learned kineme patterns, we implemented three separate configurations to discover kinemes from the low-depressed/healthy class. For the AVEC configuration (AVEC-conf), we initially discovered 16 kinemes using the head motion data acquired from low-depressed videos. Subsequently, head pose segments from the severely depressed class plus the entire Blackdog and Pitt data were represented via the learned AVEC kinemes. An identical process was followed for the Blackdog and Pitt datasets resulting in Blackdog-conf and Pitt-conf respectively.  Tables~\ref{tab:Cls_res}--\ref{tab:Reg_gen_res} present  results for varied train-test combinations in the three configurations.
%

%
\subsection{Implementation Details:}\label{Sec:Imp} 
We evaluated generalisability based on two approaches:  
\begin{itemize}
    \item \textbf{\textit{k}-fold cross-validation:} On individual/combined datasets, classification/regression model performance was evaluated on implementing 5-repetitions of 10-fold cross-validation (50 runs). 
    \item \textbf{Separate train-test datasets:} Here, the model was trained using one or two datasets and tested on the remaining dataset(s) to examine kineme generalisability.   
\end{itemize}
For both methods, we performed video-level analysis by processing chunks of 60s, 75s, 90s and 120s length, with the video class label repeated for all chunks. While classification results are reported based on the majority label over all chunks, the mean error is reported for regression analysis. 
%

%
\subsection{Classification/Regression Models:}
 For video-level analysis, we derive a total of $8 \times 3$ statistical features over the three angular dimensions for each chunk. These features are standardised via $z$-normalization. We then train different machine learning models to perform classification and regression with the computed features. Among the models mentioned in~\cite{gahalawat2023explainable}, we implement the Random Forest, Support Vector Machine (SVM), and Extreme Gradient Boosting (XGB) models, and report the best results obtained in Tables~\ref{tab:Cls_res}--\ref{tab:Reg_gen_res} based on F1 and MAE values for classification and regression respectively. 

\subsection{Performance Measures:} 
We report the accuracy (Acc), weighted F1 score (F1), precision (Pre), and recall (Re) scores for binary classification to address class imbalance in the datasets. For severity estimation (regression), we report Mean Absolute Error (MAE) and Root Mean Square Error (RMSE) scores. Further, for the \textit{k}-fold cross-validation results, $\mu \pm \sigma$ values are tabulated for the optimal models; whereas, for the separate train-test method, we report scores achieved on the (other datasets') test set. Model parameters for the separate train-test method are set to those that produce the highest training accuracy. For regression analysis on the AVEC test set, we report the MAE and RMSE scores (Table~\ref{tab:AVECtest_comp}). We utilise the AVEC validation set to fine-tune model parameters.
%
%

\section{Results and Discussion}\label{sec:RnD}
%
\begin{table*}[ht]
\renewcommand{\arraystretch}{1.2}
\fontsize{7}{7}\selectfont
\caption{Cross-validation based classification results tabulating Accuracy (Acc), F1, Precision (Pr) and Recall (Re) scores (in $\mu \pm \sigma$ form).  Highest scores achieved per row are denoted in \textbf{bold}. (Abbreviations: \textbf{Av--AVEC, Bd--Blackdog, Pt--Pitt})} 
\vspace{-2mm}
\setlength{\tabcolsep}{4pt}
\centering
\begin{tabular}{|l ||cccc||cccc||cccc|} 
\hline
    \bf Dataset & \multicolumn{4}{c||}{\textbf{AVEC-conf}} & \multicolumn{4}{c||}{\textbf{Pitt-conf}} & \multicolumn{4}{c|}{\textbf{BlackDog-conf}}    \\ 
    & \textbf{Acc} & \textbf{F1} & \textbf{Pr} & \textbf{Re} & \textbf{Acc} & \textbf{F1} & \textbf{Pr} & \textbf{Re} & \textbf{Acc} & \textbf{F1} & \textbf{Pr} & \textbf{Re} \\ 
    \hline \hline
    \textbf{Av} & \textbf{0.93±0.07} & \textbf{0.93±0.07} & \textbf{0.94±0.06} & \textbf{0.93±0.07} & 0.60±0.13 & 0.60±0.13 & 0.64±0.14 & 0.60±0.13 & 0.61±0.10 & 0.61±0.10 & 0.65±0.12 & 0.61±0.10  \\ 
    \textbf{Pt} & 0.75±0.10 & 0.68±0.13 & 0.65±0.16 & 0.75±0.10 & \textbf{0.79±0.10} & \textbf{0.80±0.10} & \textbf{0.83±0.11} & \textbf{0.79±0.10} & 0.77±0.09 & 0.67±0.13 & 0.61±0.16 & 0.77±0.09  \\
    \textbf{Bd} & 0.63±0.15 & 0.63±0.15 & 0.71±0.16 & 0.63±0.15 & 0.66±0.15 & 0.66±0.15 & 0.73±0.17 & 0.66±0.15 & \textbf{0.70±0.15} & \textbf{0.69±0.16} & \textbf{0.76±0.15} & \textbf{0.70±0.15}  \\ 
    \textbf{Av + Pt} & \textbf{0.84±0.07} & \textbf{0.83±0.07} & \textbf{0.86±0.07} & \textbf{0.84±0.07} & 0.67±0.08 & 0.68±0.07 & 0.71±0.08 & 0.67±0.08 & 0.64±0.08 & 0.64±0.08 & 0.65±0.09 & 0.64±0.08 \\ 
    \textbf{Av + Bd}& \textbf{0.82±0.08} & \textbf{0.82±0.08} & \textbf{0.84±0.08} & \textbf{0.82±0.08} & 0.63±0.10 & 0.64±0.10 & 0.66±0.10 & 0.63±0.10 & 0.60±0.09 & 0.60±0.09 & 0.62±0.09 & 0.60±0.09 \\ 
    \textbf{Pt + Bd}  & 0.70±0.09 & 0.65±0.11 & 0.69±0.14 & 0.70±0.09 & \textbf{0.71±0.09} & \textbf{0.71±0.09} & \textbf{0.72±0.09} & \textbf{0.71±0.09} & 0.64±0.08 & 0.64±0.09 & 0.65±0.10 & 0.64±0.08  \\ 
    \textbf{All } & \textbf{0.78±0.07} & \textbf{0.77±0.08} & \textbf{0.80±0.07} & \textbf{0.78±0.07} & 0.68±0.07 & 0.67±0.08 & 0.70±0.08 & 0.68±0.07 & 0.61±0.09 & 0.61±0.08 & 0.62±0.08 & 0.61±0.09  \\ 
    \hline
\end{tabular} 
\label{tab:Cls_res}
\end{table*}
\begin{table*}[ht]
\renewcommand{\arraystretch}{1.2}
\fontsize{7}{7}\selectfont
\caption{Classification results for separate train-test combinations. Highest values per row denoted in \textbf{bold}.} 
\vspace{-2mm}
\centering
\begin{tabular}{|l ||cccc||cccc||cccc|} 
\hline
    \bf Dataset & \multicolumn{4}{c||}{\textbf{AVEC-conf}} & \multicolumn{4}{c||}{\textbf{Pitt-conf}} & \multicolumn{4}{c|}{\textbf{Blackdog-conf}}   \\ 
    & \textbf{Acc} & \textbf{F1} & \textbf{Pr} & \textbf{Re} & \textbf{Acc} & \textbf{F1} & \textbf{Pr} & \textbf{Re} & \textbf{Acc} & \textbf{F1} & \textbf{Pr} & \textbf{Re} \\ 
    \hline \hline
    \textbf{Train: AVEC, Test: Pitt + Blackdog} & 0.69 & 0.62 & 0.66 & 0.69 & \textbf{0.64} & \textbf{0.64} & \textbf{0.65} & \textbf{0.64} & 0.59 & 0.59 & 0.58 & 0.59 \\ 
    \textbf{Train: Pitt, Test: AVEC + Blackdog } & \textbf{0.51} & \textbf{0.50} & \textbf{0.51} & \textbf{0.51} & 0.49 & 0.40 & 0.47 & 0.49 & 0.51 & 0.44 & 0.52 & 0.51  \\
     \textbf{Train: Blackdog, Test: AVEC + Pitt} & 0.57 & 0.58 & 0.59 & 0.57 & \textbf{0.60} & \textbf{0.59} & \textbf{0.58} & \textbf{0.60} & 0.56 & 0.54 & 0.53 & 0.56 \\ 
    \textbf{Train: Pitt + Blackdog, Test: AVEC} & \textbf{0.64} & \textbf{0.64} & \textbf{0.64} & \textbf{0.64} & 0.54 & 0.53 & 0.55 & 0.54 & 0.52 & 0.48 & 0.55 & 0.52  \\ 
    \textbf{Train: AVEC + Blackdog, Test: Pitt}  & \textbf{0.68} & \textbf{0.66} & \textbf{0.64} & \textbf{0.68} & 0.62 & 0.64 & 0.68 & 0.62 & 0.60 & 0.62 & 0.66 & 0.60 \\ 
    \textbf{Train: AVEC + Pitt, Test: Blackdog} & \textbf{0.53} & \textbf{0.47} &\textbf{ 0.54} & \textbf{0.53} & 0.50 & 0.46 & 0.48 & 0.50 & 0.49 & 0.39 & 0.42 & 0.49 \\ 
    \hline
\end{tabular} 
\label{tab:Cls_gen_res}
\end{table*}
\begin{table*}[ht]
\renewcommand{\arraystretch}{1.2}
\fontsize{7}{7}\selectfont
\caption{Cross-validation based regression results for different dataset combinations. MAE and RMSE are tabulated as ($\mu \pm \sigma$) values with the lowest values achieved per row denoted in \textbf{bold}.}
\vspace{-2mm}
\centering
\begin{tabular}{|l||cc||cc||cc|} 
\hline
    \bf Dataset    & \multicolumn{2}{c||}{\textbf{AVEC-conf}} & \multicolumn{2}{c||}{\textbf{Pitt-conf}} & \multicolumn{2}{c|}{\textbf{Blackdog-conf}} \\ 
    & \textbf{MAE} & \textbf{RMSE}  & \textbf{MAE} & \textbf{RMSE}  & \textbf{MAE} & \textbf{RMSE}  \\ 
    \hline \hline
    \textbf{AVEC} &  \textbf{3.16±0.56} & \textbf{3.93±0.64} & 5.06±0.82 & 6.00±0.82 & 5.19±0.75 & 5.97±0.71 \\ 
    \textbf{Pitt} & 5.99±0.88 & 7.02±0.93 & \textbf{5.65±0.93} & \textbf{6.63±0.96} & 6.08±0.81 & 7.02±0.82 \\
    \textbf{Blackdog} & 7.67±1.25 & 8.70±1.36 & 7.15±1.57 & 8.52±1.49 & \textbf{6.48±1.32} & \textbf{7.62±1.37} \\ 
    \textbf{AVEC + Pitt} & \textbf{4.67±0.54} & \textbf{5.76±0.59} & 5.51±0.72 & 6.44±0.76  & 5.61±0.59 & 6.54±0.56\\ 
    \textbf{AVEC + Blackdog}& \textbf{4.98±0.76} & \textbf{6.31±0.91} & 5.87±0.73 & 6.96±0.77  & 6.01±0.78 & 7.08±0.84\\ 
    \textbf{Pitt + Blackdog}  & 6.82±0.77 & 7.79±0.76 & \textbf{6.62±0.73} & \textbf{7.83±0.91}  & 6.65±0.82 & 7.79±0.82\\ 
    \textbf{All three} & \textbf{5.48±0.62} & \textbf{6.78±0.67} & 5.99±0.56 & 6.99±0.55 & 6.20±0.56 & 7.29±0.59 \\ 
    \hline
\end{tabular} 
\label{tab:Reg_res}
\end{table*}
\begin{table*}[ht]
\renewcommand{\arraystretch}{1.2}
\fontsize{7}{7}\selectfont
\caption{Regression results for separate train-test combinations. Lowest RMSE/MAE values per row are denoted in \textbf{bold}.} \vspace{-2mm}
\centering
\begin{tabular}{|l ||cc||cc||cc|} 
\hline
    \bf Dataset & \multicolumn{2}{c||}{\textbf{AVEC-conf}} & \multicolumn{2}{c||}{\textbf{Pitt-conf}} & \multicolumn{2}{c|}{\textbf{Blackdog-conf}} \\ 
    & \textbf{MAE} & \textbf{RMSE} & \textbf{MAE} & \textbf{RMSE}  & \textbf{MAE} & \textbf{RMSE}  \\ 
    \hline \hline
    \textbf{Train: AVEC, Test: Pitt + Blackdog} & \textbf{7.09} & \textbf{8.20} & 7.20 & 8.24  & 7.40 & 8.45 \\ 
    \textbf{Train: Pitt, Test: AVEC + Blackdog } & 6.85 & 8.06 & 7.30 & 8.83  & \textbf{6.79} & \textbf{7.93}\\
    \textbf{Train: Blackdog, Test: AVEC + Pitt} & \textbf{5.84} & \textbf{7.06} & 5.96 & 7.26 & 6.03 & 7.25 \\ 
    \textbf{Train: Pitt + Blackdog, Test: AVEC} & \textbf{4.75} & \textbf{5.91} & 5.26 & 6.07 & 5.45 & 6.28 \\ 
    \textbf{Train: AVEC + Blackdog, Test: Pitt}  & \textbf{6.68} & \textbf{7.91} & 6.82 & 8.19  & 6.96 & 8.55\\ 
    \textbf{Train: AVEC + Pitt, Test: Blackdog} & 8.30 & 9.34 & \textbf{8.13} & \textbf{9.01} & 8.30 & 9.38 \\ 
    \hline
\end{tabular} 
\label{tab:Reg_gen_res}
\end{table*}
\begin{table*}[ht]
\renewcommand{\arraystretch}{1.2}
\fontsize{7}{7}\selectfont
\caption{Comparison with previous classification-based generalisability results.}
\vspace{-2mm}
\centering
\begin{tabular}{|l||cc cc||c|} 
    \hline
    \bf Dataset  & \multicolumn{4}{c||}{\textbf{Ours}} &
    {\textbf{SVC~\cite{alghowinem2015cross}}}    \\ 
     & \textbf{Acc} & \textbf{F1 Score} & \textbf{Precision} & \textbf{Recall} & \textbf{Average Recall} \\ 
    \hline\hline
   \textbf{AVEC} & \textbf{0.93} & \textbf{0.93} & \textbf{0.93} & \textbf{0.93} & 0.72 \\ 
    \textbf{Pitt} & 0.79 & 0.80 & 0.83 & 0.79 & \textbf{0.87} \\ 
    \textbf{Blackdog} & 0.70 & 0.69 & 0.76 & 0.70 &\textbf{ 0.73 } \\ 
    \textbf{AVEC + Pitt} & \textbf{0.84} & \textbf{0.83} & \textbf{0.86} & \textbf{0.84} & 0.66 \\ 
    \textbf{AVEC + Blackdog} & \textbf{0.82} & \textbf{0.82} & \textbf{0.84} & \textbf{0.82} & 0.67 \\ 
    \textbf{Pitt + Blackdog} & \textbf{0.71} & \textbf{0.71} & \textbf{0.72} & \textbf{0.71} & 0.68 \\ 
    \textbf{All three} & \textbf{0.78} & \textbf{0.77} & \textbf{0.80} & \textbf{0.78} & 0.66 \\ 
    \hline
    \hline
    \textbf{Train: AVEC, Test: Pitt + Blackdog} & \textbf{0.64} & \textbf{0.64} & \textbf{0.65} & \textbf{0.64} & 0.43 \\ 
    \textbf{Train: Pitt, Test: AVEC + Blackdog} & \textbf{0.51} & \textbf{0.50} & \textbf{0.51} & \textbf{0.51} & 0.42 \\ 
    \textbf{Train: Blackdog, Test: AVEC + Pitt} & \textbf{0.60} & \textbf{0.59} & \textbf{0.58} & \textbf{0.60} & 0.43  \\ 
    \textbf{Train: Pitt + Blackdog, Test: AVEC} & \textbf{0.64} & \textbf{0.64} & \textbf{0.64} & \textbf{0.64} & 0.34 \\ 
    \textbf{Train: AVEC + Blackdog, Test: Pitt} & \textbf{0.68} & \textbf{0.66} & \textbf{0.64} & \textbf{0.68} & 0.29 \\
    \textbf{Train: AVEC + Pitt, Test: Blackdog} & 0.53 & 0.47 & 0.54 & 0.53 & \textbf{0.62} \\ 
    \hline
\end{tabular}
\label{tab:Cls_comp}
\end{table*}
\begin{table*}[ht]
\renewcommand{\arraystretch}{1.2}
\fontsize{7}{7}\selectfont
\caption{Comparison with previous regression-based generalisability results.}
\vspace{-2mm}
\centering
\begin{tabular}{|l||cc||cc||cc||cc||cc|} 
    \hline
    \bf Dataset  & \multicolumn{2}{c||}{\textbf{Ours}} & \multicolumn{2}{c||}{\textbf{AlexNet}~\cite{ahmad2021cnn}} & \multicolumn{2}{c||}{\textbf{DenseNet}~\cite{ahmad2021cnn}} & 
    \multicolumn{2}{c||}{\textbf{ResNet}~\cite{ahmad2021cnn}} &
    \multicolumn{2}{c|}{\textbf{VGG19}~\cite{ahmad2021cnn}}    \\ 
    & \textbf{MAE} & \textbf{RMSE} & \textbf{MAE} & \textbf{RMSE} & \textbf{MAE} & \textbf{RMSE} & \textbf{MAE} & \textbf{RMSE} & \textbf{MAE} & \textbf{RMSE}\\ 
    \hline \hline
    \textbf{AVEC}   & \textbf{3.16} & \textbf{3.93} & 3.98 & 4.76 & 3.63 & 4.59 & 3.77 & 4.63 & 3.80 & 4.64 \\ 
    \textbf{Blackdog}& \textbf{ 6.48 }& \textbf{7.61} & 9.53 & 10.66 & 8.32 & 10.63 & 8.32 & 10.29 & 8.48 & 9.86  \\
    \textbf{AVEC + Blackdog} & \textbf{4.96 }& \textbf{6.30} & 6.06 & 7.39 & 5.74 & 7.84 & 5.71 & 7.20 & 5.58 & 6.99  \\ \hline
    \hline
    \textbf{Train: Blackdog, Test: AVEC} & \textbf{4.90} & \textbf{5.79} & 5.32 & 6.54 & 6.23 & 7.80 & 5.84 & 7.09 & 6.12 & 7.54 \\ 
    \textbf{Train: AVEC, Test: Blackdog} & \textbf{7.92} & \textbf{8.86} & 9.38 & 9.80 & 9.30 & 9.81 & 9.52 & 10.05 & 9.30 & 9.83 \\ 
    \hline
\end{tabular}
\label{tab:Reg_comp}
\end{table*}
\begin{table*}[ht]
\renewcommand{\arraystretch}{1.2}
\fontsize{7}{7}\selectfont
\caption{Comparison with prior works on AVEC2013 test set. Best results are denoted in \textcolor{red}{red}, and our results in \textbf{bold} font.} \vspace{-2mm}
\centering
\begin{tabular}{|l l l||cc|} 
\hline
    \bf Reference & \bf Features   & \bf Methods    & \multicolumn{2}{c|}{\textbf{Evaluation metrics}}     \\ 
     & & &\textbf{MAE} & \textbf{RMSE} \\ 
    \hline\hline
    \textbf{Baseline~\cite{valstar2013avec}} & Head and facial appearance features & Support Vector Regression &  10.88 & 13.61  \\ 
    \textbf{Wen~\etal~\cite{wen2015automated}} & Sequential facial regions & Support Vector Regression &  8.22 & 10.27 \\
    \textbf{Zhu~\etal~\cite{zhu2017automated}} & Facial appearance and temporal features & Deep Convolutional Neural Network & 7.58 & 9.82 \\
    \textbf{Zhou~\etal~\cite{zhou2018visually}} & Sequential facial regions & Deep Regression Network & 6.20 & 8.28  \\
    \textbf{Li~\etal~\cite{li2020depression}} & Enhanced facial images & Deep Residual Regression Convolutional Neural Network  &  6.18 & 8.08 \\ 
    \textbf{Melo~\etal~\cite{de2020deep}} & Frontal facial appearance and temporal features & Multiscale Spatiotemporal Network &  5.98 & 7.90  \\
    \textbf{He~\etal~\cite{he2021automatic}} & Local facial patches and full facial images & Deep Local Global Attention CNN & 6.59 & 8.39 \\
    \textbf{Uddin~\etal~\cite{uddin2022deep}} & Facial appearance and temporal features & Spatio-temporal Network &  5.90 & 7.32  \\
    \textbf{Bargshady~\etal~\cite{bargshady2023estimating}}  & Multi-scale spatio-temporal facial features & Ensemble Model - SGCNN and Conv3d & 6.09 & 8.05\\
    \textbf{Pan~\etal~\cite{pan2023integrating}} & Facial landmark features & Spatial-Temporal Attention Network &  5.97 & 7.26 \\
    \textbf{Liu~\etal~\cite{liu2023net}} & Local and global facial features & Part-and-Relation Attention Network &  6.08 & 7.59    \\ 
    \textbf{Xu~\etal~\cite{xu2024two}} & Facial behavioural features & Spectral Graph Representation &  5.95 & 7.57  \\
    \textbf{Pan~\etal~\cite{pan2024spatial}} & Local and global spatio-temporal facial images & Spatial–Temporal Attention Depression Recognition Network &  6.15 & 7.98  \\    
    \textbf{Niu~\etal~\cite{niu2024depressionmlp}} & Facial keypoints and action units & Multi-Layer Perceptrons with gating &  \textcolor{red}{5.43} & \textcolor{red}{7.49}  \\
    \textbf{Song~\etal~\cite{song2020spectral}} & Head Pose descriptors &  Convolution Neural Network & 8.54 & 10.38 \\
    \textbf{Ours} & \textbf{Kinemes} & \textbf{XGBoost Regressor} &  \textbf{5.68} & \textbf{7.57} \\  \hline
\end{tabular} 
\label{tab:AVECtest_comp}
\end{table*}
Tables~\ref{tab:Cls_res} and~\ref{tab:Cls_gen_res} respectively present \textit{k}-fold and separate train-test classification results for the AVEC, Pitt and Blackdog configurations. Depression severity estimation results are tabulated in Tables~\ref{tab:Reg_res} and~\ref{tab:Reg_gen_res}. Video-level results are compiled based on a chunk size of $n_c$ = 60s for both classification and regression. Notable insights are summarised below: 
\begin{itemize}
    \item Considering Tables~\ref{tab:Cls_res} and ~\ref{tab:Reg_res}, it can be observed all models achieve the best classification/regression results when trained and tested on the same dataset, implying that kinemes are most effective at characterizing same-distribution data. 
    \item Despite the higher risk of overfitting with the separate train-test method, multiple combinations achieve reasonably above-chance-level performance (Table~\ref{tab:Cls_gen_res}), including models (i) trained on AVEC data and tested on combination of Pitt and Blackdog, (ii) trained on Blackdog and Pitt dataset and tested on AVEC, and (iii) trained on combination of Blackdog and AVEC data and tested on Pitt dataset. 
    \item Two combinations having poor classification performance include training on Pitt and testing on the combination of AVEC+Blackdog, and training on AVEC+Pitt with testing on Blackdog. The latter achieves the lowest F1 score among three configurations involving a heterogeneous training set (Table~\ref{tab:Cls_gen_res}). Poor performance obtained while testing on the Blackdog dataset could be attributed to the design differences among the three datasets. While the AVEC and Pitt datasets  classify subjects as \textit{low depressed} and \textit{severely depressed},  Blackdog categories comprise \textit{healthy controls} and \textit{depressed} individuals, which impairs feature generalisability.
    \item For separate train-test regression experiments (Table~\ref{tab:Reg_gen_res}),  all three heterogeneous train sets (rows 4--6) achieve reasonably good performance. Comparing the best MAE scores from different train-test dataset combinations in Table~\ref{tab:Reg_gen_res} reveals that optimal results on AVEC are achieved with training on Pitt+Blackdog (MAE/RMSE:  4.75/5.91) as compared to a model trained solely on Blackdog (MAE/RMSE - 4.90/5.79). Conversely, optimal results on Blackdog are achieved with a model trained on AVEC only (Table~\ref{tab:Reg_comp}, MAE/RMSE - 7.92/8.86). This indicates larger range and variance in the AVEC head movements, necessitating heterogeneous training features, as compared to Blackdog. 
    \item Consistent with the above, kinemes discovered from  AVEC (AVEC-conf) generally achieve better F1/MAE scores as compared to the other configurations. This is evident from the classification performance on all three datasets (F1=0.77, Table~\ref{tab:Cls_gen_res}) achieved via kinemes learned from AVEC dataset versus F1 scores of 0.61 and 0.67, respectively, with Blackdog-conf and Pitt-conf. We notice a similar trend for regression results in Table~\ref{tab:Reg_res}. These results demonstrate a stronger generalisation power of AVEC kinemes, which can be attributed to data capture context-- while both Blackdog and Pitt datasets were recorded in an interview setting, AVEC participants self-completed a set of tasks. Absence of an interviewer enables AVEC subjects to express themselves more naturally, leading to distinctive head movement behaviors for the low and high depressed cohorts.
    \item Additionally, AVEC-conf achieves superior performance on dataset combinations (Table~\ref{tab:Cls_res}), compared to Blackdog-conf and Pitt-conf. Regression results in Table~\ref{tab:Reg_res} follow a consistent pattern with AVEC-conf achieving best performance on heterogeneous test sets, confirming the efficacy of AVEC-derived kinemes.  
\end{itemize}

%
%
\subsection{Comparison with Prior Classification Approaches}
For binary classification, Table~\ref{tab:Cls_comp} compares our best results with Alghowinem~\etal~\cite{alghowinem2015cross}, who investigated head pose and eye gaze as depression biomarkers along with their fusion. To ensure a fair comparison, we compare our results with their best results obtained with both fixed and variable head-pose features. Experiments in Alghowinem~\etal~\cite{alghowinem2015cross} involved an AVEC subset of 16 subjects per class, a balanced subset of 60 videos for Blackdog, and 19 participants per class for the Pitt dataset. The authors employed \emph{Constrained Local Models} for detecting fiducial face points, which were projected onto a 3D face model to extract head pose and 184 statistical features therefrom. These features were used to train an SVM classifier, which was evaluated on individual as well as combinations of datasets. Considering individual dataset performance, kinemes achieve substantially better performance on AVEC, comparable performance on Blackdog and inferior performance on the Pitt dataset. Kinemes outperform raw head pose features in dataset combinations (rows 5--7), highlighting the generalisation efficacy of kinemes over varied observations.  For separate train-test combinations (rows 8--13), a better performance can be observed for all-but-one cases with the kineme approach. 

%
\subsection{Comparison with prior regression algorithms}
Table~\ref{tab:Reg_comp} compares our best results against prior depression severity estimation studies. As only a few studies have explored depression estimation generalisability, we compare our results with Ahmad~\etal~\cite{ahmad2021cnn} employing convolution neural networks (CNNs). which utilized the AVEC train and development sets and a gender-and-age balanced subset of the Blackdog dataset. Multiple CNN models such as AlexNet~\cite{krizhevsky2012imagenet}, VGG~\cite{simonyan2014very}, ResNet~\cite{he2016deep} and DenseNet~\cite{huang2017densely} on facial features are extracted from video frames, and 5-fold cross-validation on individual and combination of datasets is employed for evaluation. To ensure fair comparison, we also employed evaluation settings identical to~\cite{ahmad2021cnn} for evaluation and present the best results. From the table, we observe that kinemes outperformfacial features for individual datasets as well as their combinations, with a substantial difference for Blackdog (MAE of 6.48 vs 8.32). A similar performance improvement is noticed for separate train-test combinations, confirming generalisability effectiveness of kineme patterns for depression severity estimation.

%
\subsection{Comparison with AVEC-based regression studies}
To showcase the efficacy of kinemes for depression severity estimation, we compare our approach against other visual methodologies on the AVEC test set in Table~\ref{tab:AVECtest_comp}.  To our knowledge, only Song~\etal~\cite{song2020spectral} have investigated depression analysis using head movements for AVEC; the authors used head pose-based spectral representations, which are then fed to CNNs for depression estimation. Kinemes achieve substantially better performance than~\cite{song2020spectral} (MAE of 5.68 vs 8.54). Table~\ref{tab:AVECtest_comp} demonstrates that our kineme-based approach outperforms state-of-the-art visual approaches, excepting~\cite{niu2024depressionmlp} where the authors utilise facial key-points and action units for depression estimation.

%
%
\section{Conclusion}\label{Sec:Con}
This paper illustrates the effectiveness of fundamental head motion patterns termed kinemes for estimating depression severity. Utilising three cross-cultural datasets with varied contextual settings, our results demonstrate enhanced generalisability of kineme-based features versus (a) raw head pose features for binary classification, and (b) other visual cues for severity estimation. Future work involves (a) improving the kineme discovery framework patterns in a bid to enhance their interpretability, and (b) developing kineme-based multimodal approaches for depression assessment. 

\section*{Acknowledgement}
This research is partially funded by the Australian Government through the Australian Research Council’s Discovery Projects funding scheme (Project DP190101294) and US National Institutes of Health under Award Number MH096951. Monika Gahalawat is supported by a scholarship from the Faculty of Science \& Technology (University of Canberra).

\bibliographystyle{IEEEtran}
\bibliography{references}

\end{document}